\documentclass[11pt]{article}

\usepackage[preprint]{acl}

\usepackage{times}
\usepackage{latexsym}

\usepackage[T1]{fontenc}
\usepackage{amsmath}
\usepackage{amssymb}
\usepackage[utf8]{inputenc}

\usepackage{microtype}

\usepackage{inconsolata}

\usepackage{graphicx}

\usepackage{xcolor}

\usepackage{enumitem} 
\setenumerate[1]{itemsep=0pt,partopsep=0pt,parsep=\parskip,topsep=5pt}
\setitemize[1]{itemsep=0pt,partopsep=0pt,parsep=\parskip,topsep=5pt}
\setdescription{itemsep=0pt,partopsep=0pt,parsep=\parskip,topsep=5pt}

\usepackage{scalerel} 
\usepackage{booktabs} 
\usepackage{pdfpages}
\usepackage{multirow}
\usepackage{subfig}
\usepackage[capitalize]{cleveref}
\usepackage{xspace}
\usepackage{cleveref}
\usepackage{url}

\newcommand{\method}
{\textsc{HybridKV}\xspace}

\usepackage{amssymb}  
\usepackage{pifont}

\usepackage[capitalize]{cleveref}
\definecolor{c0}{cmyk}{1,0.3968,0,0.2588} 
\definecolor{LightCyan}{rgb}{0.88,1,1}
\newcommand{\gray}[1]{\cellcolor{gray!10}\textcolor{gray}{#1}} 
\usepackage{scalerel} 

\usepackage[normalem]{ulem}
\usepackage{makecell} 
\usepackage{colortbl} 

\usepackage{scalerel} 

\definecolor{uclablue}{rgb}{0.15, 0.45, 0.68}
\definecolor{custompink}{RGB}{233,0,136}

\hypersetup{
    breaklinks,
    citecolor=uclablue,
    colorlinks=true,
    linkcolor=uclablue,
    urlcolor=custompink
}

\title{
\scalerel*{\includegraphics{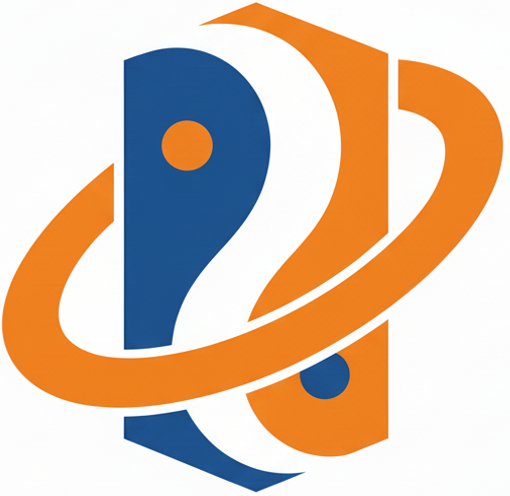}}{{\rule{2.2ex}{2.2ex}}}
\method: Hybrid KV Cache Compression for Efficient Multimodal Large Language Model Inference}

\author{
  \textbf{Bowen Zeng\textsuperscript{1,2\ding{44}}},
  \textbf{Feiyang Ren\textsuperscript{1,2\ding{44}}},  
  \textbf{Jun Zhang\textsuperscript{1,2\ding{44}}}, 
  \textbf{Xiaoling Gu\textsuperscript{3}},\\
  \textbf{Ke Chen\textsuperscript{1,2},} 
  \textbf{Lidan Shou\textsuperscript{1,2},}
  \textbf{Huan Li\textsuperscript{1,2\ding{41}}}\\
  \textsuperscript{1}The State Key Laboratory of Blockchain and Data Security, Zhejiang University \\
  \textsuperscript{2}Hangzhou High-Tech Zone (Binjiang) Institute of Blockchain and Data Security \\
  \textsuperscript{3}Hangzhou Dianzi University, Hangzhou, China \\
  \small{\tt\{zbw.cs, feiyangren, zj.cs, chenk, should, lihuan.cs\}@zju.edu.cn, guxl@hdu.edu.cn} 
}

\begin{document}
\maketitle

\let\oldthefootnote\thefootnote
\renewcommand{\thefootnote}{}

\footnotemark
\footnotetext{\textsuperscript{\ding{44}}Equal contribution. \textsuperscript{\ding{41}}Corresponding author.}

\let\thefootnote\oldthefootnote

\begin{abstract}
Multimodal Large Language Models (MLLMs) have advanced unified reasoning over text, images, and videos, but their inference is hindered by the rapid growth of key–value (KV) caches.
Each visual input expands into thousands of tokens, causing caches to scale linearly with context length and remain resident in GPU memory throughout decoding, which leads to prohibitive memory overhead and latency even on high-end GPUs.
A common solution is to compress caches under a fixed allocated budget at different granularities: token-level uniformly discards less important tokens, layer-level varies retention across layers, and head-level redistributes budgets across heads. Yet these approaches stop at allocation and overlook the heterogeneous behaviors of attention heads that require distinct compression strategies.
We propose \method, a hybrid KV cache compression framework that integrates complementary strategies in three stages: heads are first classified into static or dynamic types using text-centric attention; then a top-down budget allocation scheme hierarchically assigns KV budgets; finally, static heads are compressed by text-prior pruning and dynamic heads by chunk-wise retrieval.
Experiments on 11 multimodal benchmarks with \texttt{Qwen2.5-VL-7B} show that \method reduces KV cache memory by up to $7.9\times$ and achieves $1.52\times$ faster decoding, with almost no performance drop or even higher relative to the full-cache MLLM.

\end{abstract}
\section{Introduction}
\label{sec:intro}


Multimodal Large Language Models (MLLMs) such as \texttt{Qwen2.5-VL}~\cite{bai2025qwen2}, \texttt{LLaVA-OneVision}~\cite{li2024llava}, and \texttt{InternVL}~\cite{chen2024internvl} mark a major step forward in extending large language models beyond text, enabling unified reasoning over both language and visual modalities.
To support such capability, they process long multimodal contexts in which each high-resolution image or video frame unfolds into hundreds of tokens, rapidly inflating sequence lengths.
This expansion directly drives a linear growth of Key–Value (KV) caches, which store the key and value representations of all past tokens for attention computation and must be retained throughout autoregressive decoding.
For instance, a 72B-scale \texttt{Qwen2.5-VL} processing 20 images already exceeds 40K tokens and 13 GB of cache, while a 5-second 720p video surpasses 50K tokens and 16 GB. 
Even a small batch of five inputs nearly saturates an entire 80 GB A100 GPU.
Such oversized caches impose heavy GPU memory demands and incur substantial latency from repeated memory access during decoding, ultimately limiting the practical use of MLLMs.



\begin{figure} [!t] 
\centering
    \includegraphics[scale=0.37]{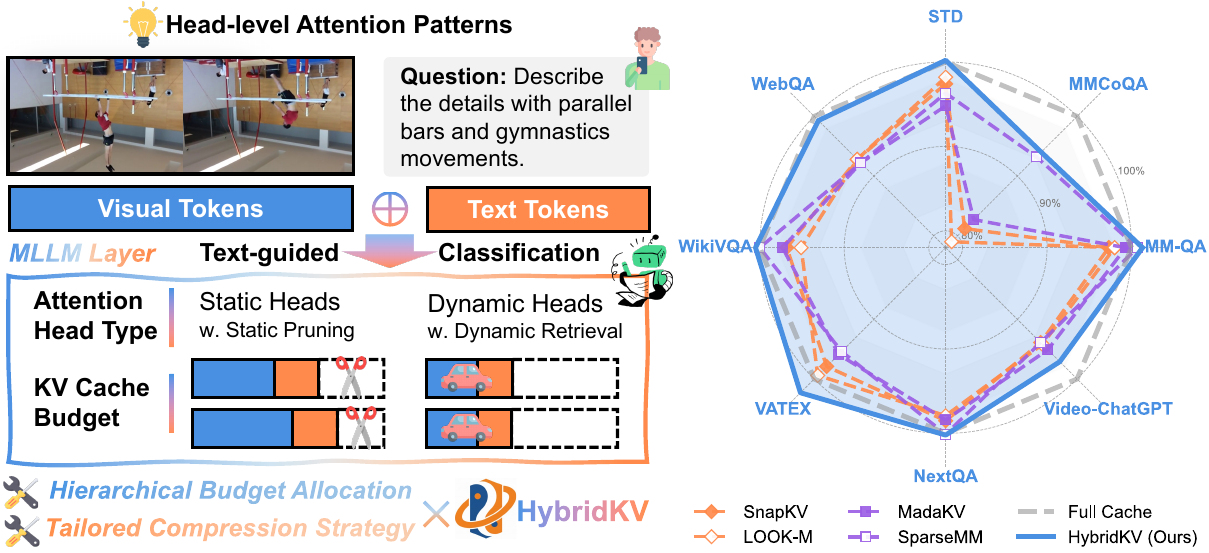}
    \caption{\textbf{Left}: We introduce \method, a hybrid KV cache compression framework for efficient yet effective MLLM inference. 
    \method leverages head-level attention patterns (detailed in~\cref{sec:obs}) to classify static and dynamic heads via text-guided signals, enabling hierarchical budget allocation with tailored \emph{pruning} and \emph{retrieval} strategies. 
    \textbf{Right}: \method outperforms existing counterparts including \textsc{SnapKV}~\cite{li2024snapkv}, \textsc{LOOK-M}~\cite{wan2024look}, \textsc{MadaKV}~\cite{li2025madakv} and \textsc{SparseMM}~\cite{wang2025sparsemm} on eight benchmarks using only 10\% of the KV cache on \texttt{Qwen2.5-VL-7B}.
    Performance metrics are shown as a percentage relative to \textsc{Full Cache}.
    }
    \label{fig:intro_map2}
\end{figure}


To alleviate these memory and latency bottlenecks, several KV cache compression methods have been proposed for MLLM inference. Under a restricted KV cache budget, these methods aim to optimize budget allocation to achieve generation quality comparable to using the full cache.
They can be broadly categorized into token-level, layer-level, and head-level strategies that target different granularities of budget allocation. 
Token-level methods (e.g.,~\citet{wan2024look,li2024snapkv,xiao2024efficient}) allocate this budget uniformly across layers and heads, discarding unimportant tokens. Layer-level methods (e.g.,~\citet{li2025madakv,wan2025meda,cai2024pyramidkv}) instead vary the allocation across layers, motivated by the intuition that different layers exhibit different sensitivities to compression amounts, so some layers can be pruned more aggressively while others require more tokens to be preserved. Head-level methods (e.g.,~\citet{wang2025sparsemm,feng2024ada}) operate at the finest granularity, guided by the intuition that different attention heads capture distinct features or modalities, leading to highly imbalanced importance across heads.
While head-level methods push compression to the finest granularity by redistributing budgets across heads, they remain confined to deciding how much each head should be pruned.
Yet performance gaps persist, as certain heads tolerate aggressive eviction while others collapse even under mild pruning.
This naturally raises a deeper question: 

\textit{Should accurate KV cache compression stop at budget allocation, or should it also involve \textbf{strategies tailored to different heads}?}

We argue that accurate KV cache compression requires both budget allocation and strategies tailored to different heads.
Allocating budgets is necessary but insufficient, because attention heads exhibit distinct behaviors across different stages of inference.
We observe that some MLLM heads follow a static pattern, where their KV entries remain stable once established during prefilling and can be pruned safely.
Others display a dynamic pattern, where important entries continue to evolve during decoding and cannot be pruned without severe degradation (\cref{sec:obs:head_pattern}).
This motivates \method, a hybrid compression framework as shown in~\cref{fig:intro_map2} (left). 
\method offers mechanisms to differentiate behaviors of attention heads, and assigns each category a tailored compression method with properly allocated budgets.
It operates in three stages.
First, since MLLMs interpret visual inputs under textual guidance, we use text tokens as stable anchors and compute a text-centric sparsity score to classify heads into static and dynamic types (\cref{sec:head-classify}).
Second, to account for the heterogeneous sensitivity of these heads to compression, \method proposes a top-down budget allocation strategy that hierarchically allocates KV capacity for both different head types and each individual head (\cref{section_3.2}).
Finally, static heads are compressed with \emph{text-prior pruning}, which retains salient KV entries guided by the attention of local window tokens, while dynamic heads are handled with \emph{chunk-wise retrieval}, which stores their KV cache in blocks and selectively retrieves important chunks during decoding (\cref{sec:met:hybridkv_impl}).
Together, these stages enable efficient and accurate multimodal inference, outperforming existing KV cache compression methods on \texttt{Qwen2.5-VL-7B} across image and video benchmarks (see~\cref{fig:intro_map2} (right)).

To summarize, our main contributions are:

\begin{itemize}[itemsep=1.5pt, topsep=1.5pt, leftmargin=20pt]
    \item[(1)] \textit{Novel Pattern Discovery} (\cref{sec:obs}): We perform a systematic analysis of head-level properties within MLLMs, revealing how hybrid attention patterns can guide head-level KV cache compression.
    \item[(2)] \textit{Hybrid Compression Scheme} (\cref{sec:met}): We propose \method, an efficient MLLM inference framework that uses hierarchical KV cache allocation and head-specific compression guided by modality-aware features, enabling superior KV cache compression.
    \item[(3)] \textit{Extensive Experiments} (\cref{sec:exp}): Evaluations on both image and video benchmarks show that \method compresses 90\% of KV caches with minimal accuracy loss and delivers up to $1.52\times$ faster decoding on \texttt{Qwen2.5-VL-7B}.
\end{itemize}

\section{Observation}
\label{sec:obs}

\begin{figure*} [!t] 
\centering
    \includegraphics[scale=0.4]{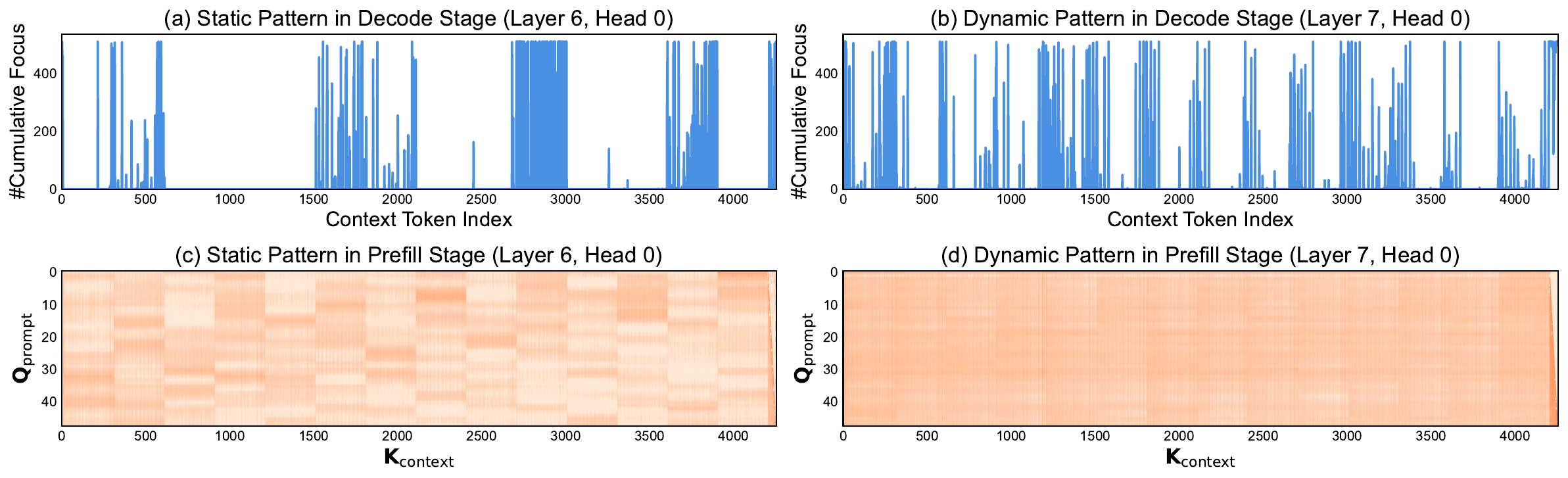}
    \caption{Static vs. dynamic attention patterns in MLLM inference. The decode stage (\emph{a}, \emph{b}) strongly correlates with the corresponding prefill stage (\emph{c}, \emph{d}). \textbf{(a, c) Static Heads}: focus stably on a small set of tokens during decoding (\emph{a}), consistent with high‑sparsity text‑centric attention (\cref{eq:score_matrix}) in prefill (\emph{c}). \textbf{(b, d) Dynamic Heads}: shift focus across many tokens during decoding (\emph{b}), consistent with low‑sparsity distributions in prefill (\emph{d}).}
    \label{fig:obs_head_patttern}
\end{figure*}

\subsection{Head-level Attention Patterns}
\label{sec:obs:head_pattern}

To examine whether distinct patterns exist among attention heads in MLLMs, we conduct a pilot study on \texttt{Qwen2.5-VL-7B} using the \textsf{VATEX}~\cite{wang2019vatex} benchmark\footnote{\textsf{VATEX} provides richly annotated video–text pairs that expand into long multimodal contexts, making it an ideal testbed when combined with a relatively large 7B MLLM.}. 
The measure is implemented with a \textbf{\emph{cumulative focus count}}.
For each context token, this metric counts how often it appears among the top‑$k$ attended tokens across multiple decoding steps. Formally, for a specific head, the cumulative focus for the $j$-th token is
\begin{equation}
    {F_j} = \sum\nolimits_{i=1}^{N} \mathbb{I}(j \in \text{TopK}_{\text{indices}}(\mathbf{{a}}^{(i)},k)),
    \label{eq:focus_count}
\end{equation}
where $N$ is the number of decoding steps, $\mathbf{a}^{(i)}$ is the attention score vector over the context at step $i$, $\mathbb{I}(\cdot)$ is the indicator function for counting, and $k = \lceil0.05\times C\rceil$ with $C$ the context length.
We set $N = 500$ to track behavior across a long generation.

Analyzing the cumulative focus counts reveals two clear patterns:
(i) \textbf{static pattern}, a head repeatedly attends to a specific and limited set of context tokens (see~\cref{fig:obs_head_patttern} (a)), indicating stable focus; and
(ii) \textbf{dynamic pattern}, a head distributes attention more fluidly across many tokens (see~\cref{fig:obs_head_patttern} (b)), indicating evolving focus.

To explain these head patterns, we further analyze the prefill stage, when the MLLM forms cross-modal understanding by attending from the text prompt to the entire context. 
We employ \textbf{\emph{text‑centric attention}} to quantify this interaction:
\begin{equation}
\mathbf{S_\text{text}} = \text{Softmax}\left(\frac{\mathbf{Q}_{\text{text}} \mathbf{K}_{\text{context}}^\top}{\sqrt{d_k}}\right),
\label{eq:score_matrix}
\end{equation}
where $\mathbf{Q}_{\text{text}} \in \mathbb{R}^{T \times d}$ are the vectors of query state derived from the $T$ input prompt text tokens, $\mathbf{K}_{\text{context}} \in \mathbb{R}^{C \times d}$ are the key vectors of the entire $C$ context tokens, and $d$ is the hidden state dimension. 
Examining the sparsity of the matrix $\mathbf{S}_\text{text}$, we find that heads also exhibit two distinct patterns in prefill: \textbf{high‑sparsity} (\cref{fig:obs_head_patttern} (c)), where most attention focuses on a few key tokens, and \textbf{low‑sparsity} (\cref{fig:obs_head_patttern} (d)), where attention spreads broadly.

Our analysis uncovers a \emph{clear} predictive link between prefill and decode behaviors. Heads with high-sparsity text-centric attention in the prefill stage consistently exhibit a static pattern in the decode stage (\cref{fig:obs_head_patttern} (a, c)); while those with low-sparsity attention align with the dynamic pattern (\cref{fig:obs_head_patttern} (b, d)).
%
%
The intuition is straightforward: when a head, guided by the text prompt, concentrates on a small set of tokens early on, its focus stabilizes during decoding. Conversely, a diffuse prefill distribution implies that many tokens may become relevant later, prompting the head to shift attention dynamically.
\ding{224} \textbf{\emph{Thus, the sparsity of text-centric attention in the prefill stage provides a reliable and efficient proxy for predicting whether attention heads behave statically or dynamically during decoding.}}

\begin{figure*} [!t] 
\centering
    \includegraphics[scale=0.32]{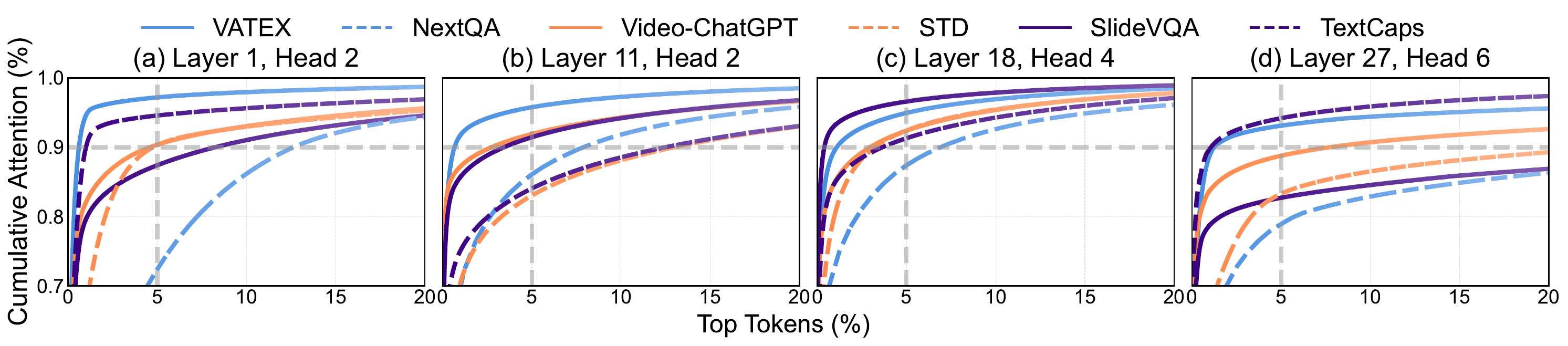}
    \caption{
    Cumulative attention distribution during prefill for selected heads in \texttt{Qwen2.5-VL-7B}. Each curve shows the fraction of attention captured by the top $x\%$ tokens across video tasks (\textsf{VATEX}, \textsf{NextQA}, \textsf{Video‑ChatGPT}) and image tasks (\textsf{STD}, \textsf{SlideVQA}, \textsf{TextCaps}).
    }
    \label{fig:obs_task_level_pattern}
\end{figure*}

\subsection{Variation of Patterns Across Tasks}
\label{sec:obs:task_pattern}
\cref{sec:obs:head_pattern} has shown that prefill sparsity predicts whether an attention head follows a static or dynamic pattern. A natural follow-up question is whether a head's such behavior is consistent across tasks or varies with context.  
To examine this, we evaluate \texttt{Qwen2.5-VL-7B} on multiple video and image benchmarks.
We use \textbf{\emph{cumulative attention distribution}} curves to quantify the head's attention sparsity during the prefill stage, as shown in \cref{fig:obs_task_level_pattern}. 
The results reveal clear \emph{task-dependent variation}. For example, in \cref{fig:obs_task_level_pattern} (a), the same head shows a highly sparse, static pattern on \textsf{VATEX} (blue solid line: top 5\% tokens cover $>95$\% of attention), but a much denser, dynamic pattern on \textsf{NextQA} (blue dashed line: top 5\% tokens cover $<75$\% of attention). \ding{224} \textbf{\emph{Thus, a head can behave statically in one task while dynamically in another.}}


Even within the static family (say, heads with $>90$\% cumulative attention), sparsity levels vary. In \cref{fig:obs_task_level_pattern} (b, c), two heads from \textsf{SlideVQA} both appear static, yet one focuses nearly 95\% of its mass on the top 5\% tokens, while the other barely crosses 90\%. A similar pattern is observed in \textsf{VATEX} (\cref{fig:obs_task_level_pattern}(a, d)).
\ding{224} \textbf{\emph{Hence, heads not only switch between static and dynamic roles across tasks but also vary in strength within the same role.}}


Since prefill sparsity governs decoding behavior, these differences confirm that a head's role is task-dependent. Effective compression must therefore be context-aware, \emph{adapting to both input and task}.

\section{\method{}}
\label{sec:met}

\begin{figure*} [!t] 
\centering
    \includegraphics[scale=0.5]{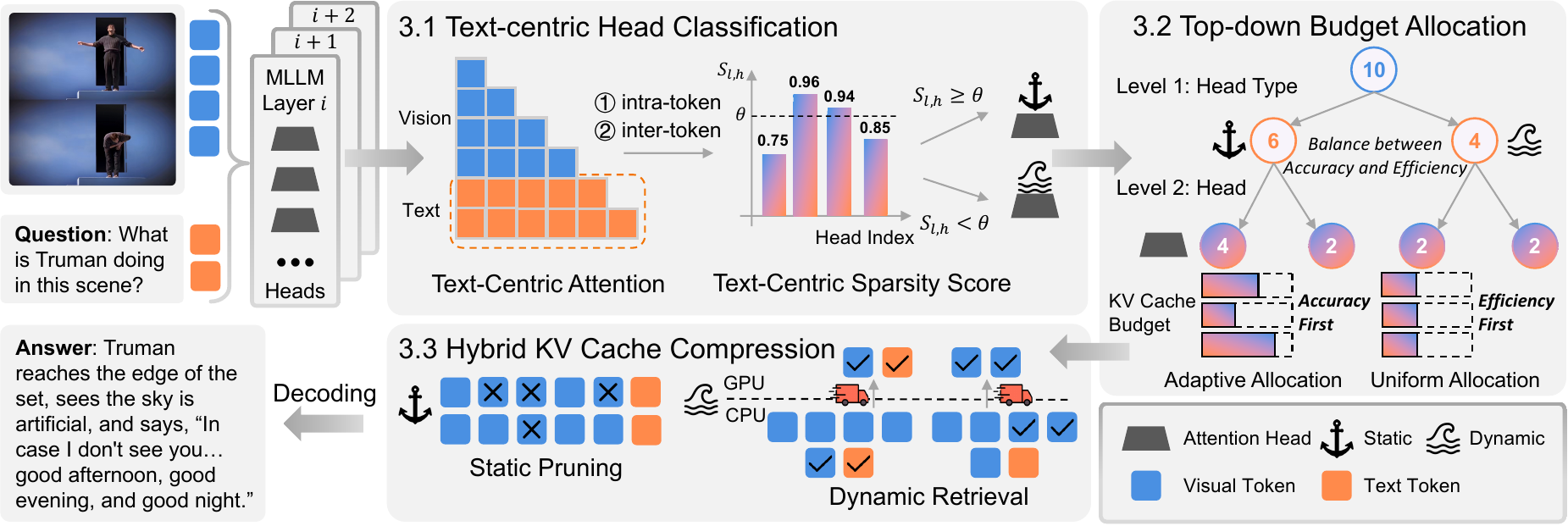}
    \caption{
    Overview of \method. \cref{sec:head-classify}: Heads are classified by text‑centric sparsity into static (anchor, few tokens) and dynamic (wave, broad focus). \cref{section_3.2}: A two-layer scheme allocates KV cache budgets first by head type and individual heads then. \cref{sec:met:hybridkv_impl}: Hybrid compression integrates static pruning (drop uninformative tokens) with dynamic retrieval (reactivate tokens when needed) to optimize cache usage.}
    \label{method_map}
\end{figure*}

Given the insights, we present \method, a hybrid KV cache compression framework for efficient yet effective MLLM inference (\cref{method_map}). 
\method exploits the heterogeneity of attention heads with a context‑aware, head‑level design. 
It first classifies heads into static or dynamic based on text‑centric sparsity (\cref{sec:head-classify}), then applies a top‑down budget allocation to distribute cache resources hierarchically (\cref{section_3.2}). 
Finally, it integrates static pruning with dynamic retrieval into a unified hybrid compression strategy (\cref{sec:met:hybridkv_impl}).

\subsection{Text-centric Head Classification}
\label{sec:head-classify}

As noted in \cref{sec:obs:head_pattern}, MLLM attention heads exhibit sparsity in text‑centric attention (\cref{eq:score_matrix}), with most attention focused on a small subset of tokens in the entire context.
Static heads show sharper focus on these tokens, while dynamic heads attend more broadly. 
This contrast motivates a principled metric for head classification.  

Building on \cref{sec:obs:task_pattern}, this metric should be \emph{context‑aware}, adapting to different inputs and tasks to ensure accurate classification across diverse conditions. To this end, we propose the \textbf{\emph{text‑centric sparsity score}}, which separates static and dynamic heads during the prefill stage. The score is computed in two steps:  
(i) \emph{intra‑token}: retain only the top‑$k$ positions from each text token's distribution to suppress long‑tail noise;  
(ii) \emph{inter‑token}: aggregate these filtered values into a head‑level profile.  
Formally, for the $h$‑th head in layer $l$ with text token set $T$:
\begin{equation}
S_{l,h} = \frac{1}{|T|} \sum\nolimits_{i=1}^{|T|} \sum\nolimits_{j=1}^{k} \text{TopK}(\mathbf{S_\text{text}}[i], k),
\end{equation}
where $\mathbf{S}_{\text{text}}[i]$ denotes the attention distribution over all tokens when attending from text token $i$, and $\text{TopK}(\cdot,k)$ selects the top-$k$ entries. 
Finally, we classify heads by thresholding $S_{l,h}$ against a quantile‑based cutoff $\theta$ tuned on a validation set. Heads with $S_{l,h} \geq \theta$ are labeled \textit{static}, while those with $S_{l,h} < \theta$ are labels \textit{dynamic} (see~\cref{appendix:param:threshold} for a sensitivity study on $\theta$).
%
Based on this classification, we can apply a hybrid compression strategy (\cref{sec:met:hybridkv_impl}):
Static heads exhibit stable attention and are compressed more aggressively via pruning, while dynamic heads adapt their focus during decoding and we use retrieval to fetch relevant tokens at runtime.
As pruning and retrieval entail different efficiency–accuracy trade‑offs, we develop a tailored budget allocation to balance the two.

\subsection{Top-down Budget Allocation}
\label{section_3.2}
Existing KV cache compression \citep{wan2024look,li2025madakv,wang2025sparsemm} for MLLMs mainly relies on uniform pruning across heads, disregarding their distinct roles and behaviors.
%
In contrast, our method leverages head classification to perform hierarchical budget allocation, first distributing \emph{budgets between head types}, then refining them \emph{across individual heads}, achieving a balanced trade-off between generation quality and efficiency.

\paragraph{Budget Allocation at the Head-type Level.}  
This step is critical as an imbalanced allocation can degrade both accuracy and efficiency. 
Over‑allocating to static heads limits the capacity to track evolving focus tokens of dynamic heads during decoding, whereas excessive allocation to dynamic heads discards important prefilling tokens and increases KV cache I/O overhead from dynamic retrieval (as depicted in \cref{sec:met:hybridkv_impl}).
%
To balance the two, we allocate the total KV budget $B_{\text{total}}$ across $N_{\text{stat}}$ static and $N_{\text{dyna}}$ dynamic heads using the average per‑head budget
$\bar{B}=B_{\text{total}}/(N_{\text{stat}}+N_{\text{dyna}})$. 
%
A share coefficient $r$, constrained by not exceeding total budget such that $r \le B_\text{total}/(\bar{B} \cdot N_\text{dyna})$, controls the portion assigned to dynamic heads:
\begin{equation}
B_\text{dyna} = \lceil r\cdot \bar{B} \cdot N_\text{dyna} \rceil; \, 
B_\text{stat} = B_\text{total}-B_\text{dyna}.
\end{equation}
The coefficient $r$ modulates the accuracy–efficiency trade‑off, with $r=1$ indicating uniform allocation for the two head types.
Its sensitivity analysis is provided in \cref{appendix:param:share_ratio}.

\paragraph{Budget Allocation at the Individual Head Level.}
After determining type‑level budgets, we further distribute them among individual heads while satisfying total memory constraints.

For static heads, whose importance varies across tasks (as observed in \cref{sec:obs:task_pattern}), we allocate budgets adaptively using the text‑centric sparsity score $S_{l,h}$ from \cref{sec:head-classify}.
%
As this score is calculated for the current input, it is inherently task-specific to quantify the importance of each head. 
The scores are normalized across all static heads with $L_1$ normalization. 
Formally, we denote the set of static heads in layer $l$ as $\mathcal{H}_{\text{stat}, l}$. If the $h$-th head in layer $l$ is in $\mathcal{H}_{\text{stat}, l}$, the budget for this static head is allocated as the sum of a uniform base budget and a score-proportional budget. We define $\alpha \in (0,1)$ as the allocation ratio, controlling the balance between the two types of budget. Formally, we compute each head's budget as
\begin{equation}
b_{l,h} = \lceil \underbrace{\alpha \cdot B_{\text{stat}} \cdot {N_{\text{stat}}^{-1}}}_{\text{Base Budget}} + \underbrace{ (1-\alpha) \cdot B_{\text{stat}} \cdot S_{l,h} }_{\text{Proportional Budget}} \rceil,
\end{equation}
This adaptive strategy ensures that important static heads receive sufficient resources, improving accuracy across diverse multimodal tasks.

For dynamic heads, to be compatible with the efficient implementation of the dynamic retrieval technique introduced in~\cref{sec:met:hybridkv_impl}, we simply allocate the total budget $B_\text{dyna}$ uniformly across $N_\text{dyna}$ dynamic heads, consistent with previous studies~\cite{tang2024quest,sun2024shadowkv}. 
The assigned budget is padded to its nearest powers of 2 to accommodate CUDA constraints in dynamic KV cache handling.
This strategy maintains decoding efficiency with minimal accuracy loss.


\subsection{Hybrid KV Cache Compression}
\label{sec:met:hybridkv_impl}
Building upon head classification and top‑down budget allocation, we implement a hybrid KV cache compression process.
For static heads, whose attention targets stable text and visual positions, we apply \textbf{\emph{text-prior static pruning}}: during prefill, KV caches for text tokens are retained with higher priority, followed by those for local and salient visual tokens, reducing memory usage.
For dynamic heads, which attend to small and variable subsets of multimodal inputs, we adopt \textbf{\emph{chunk-wise dynamic retrieval}}, widely used in recent KV cache methods~\cite{tang2024quest,sun2024shadowkv}. KV caches are offloaded to the CPU to satisfy memory constraints, and a chunk-level index is built for efficient retrieval in prefill. Important KV caches are selected dynamically for attention computation during decoding.

A unified GPU KV cache buffer accommodates important tokens from both head types, filled by static heads during prefill and updated by dynamic heads during decoding. Implementation details are provided in \cref{appendix:hybrid_kv_compress_impl}. Importantly, our hybrid compression process is orthogonal to existing static pruning and dynamic retrieval methods (see \cref{appendix:related_work}) and can be readily combined with them for further gains in accuracy and efficiency.

\section{Experiments} \label{sec:exp}

\paragraph{Baselines.}
We compare against four recent KV cache compression methods for MLLMs. \ding{172} \textsc{SnapKV}~\cite{li2024snapkv} and \ding{173} \textsc{LOOK-M}~\cite{wan2024look} allocate budgets at the token level: \textsc{SnapKV} retains crucial tokens using cumulative attention scores, while \textsc{LOOK-M} prioritizes text tokens and applies KV cache merging for MLLMs; \ding{174} \textsc{MadaKV}~\cite{li2025madakv} distributes budgets across layers based on a modality-preference metric; \ding{175} \textsc{SparseMM}~\cite{wang2025sparsemm} allocates proportional budgets to heads using offline visual head scores. All these emphasize budget allocation without considering compression strategies tailored to different head types.

\paragraph{Tasks and Models.}
We evaluate \method on both image and video tasks using MileBench~\cite{song2024milebench} and LMMs-Eval~\cite{zhang2025lmms}. Detailed benchmark information is provided in~\cref{appendix:exp_setting}. The MLLMs include \texttt{Qwen2.5-VL-3B}, \texttt{Qwen2.5-VL-7B}, and \texttt{LLaVA-OneVision-7B}
\footnote{\textcolor{black}{We exclude evaluating \texttt{InternVL} as \textsc{SparseMM} does not provide its visual head score and lacks support for it.}}. 
For video tasks, we sample 64 frames per video input. All experiments are run on NVIDIA L40S GPUs. 

\subsection{Performance Results}

The overall comparison results on the image and video tasks are reported in \cref{tab:main_result_image} and \cref{tab:main_result_video}, respectively. 
We allocate only 10\% of the full KV cache to evaluate performance under high-compression settings. Despite this strict constraint, \method maintains accuracy comparable to, and sometimes even surpasses, the \textsc{Full Cache} baseline (gray bars) on different model sizes and architectures, {yielding greater performance gains on larger models}.
These results demonstrate that \method substantially reduces the memory footprint with a negligible impact on performance. Such savings are crucial for deploying MLLMs in memory-intensive scenarios like long video understanding (e.g., \textsf{VATEX}) and multi-image reasoning (e.g., \textsf{MM-QA}), where the uncompressed KV caches exceed the capacity of even high-end GPUs.

\paragraph{Image Tasks.}
\textcolor{black}{As shown in~\cref{tab:main_result_image}, \method yields accuracy comparable to the~\textsc{Full Cache} baseline (within 1.3\% drop in average) across eight tasks 
included in MileBench~\cite{song2024milebench}, which cover complex multi-image settings.}
\textcolor{black}{
It outperforms \textsc{SnapKV}, \textsc{LOOK-M} and \textsc{MadaKV}, which rely on token- and layer-level budget allocation, across almost all MLLMs and datasets, showing the effectiveness of our head-level compression strategy.}
\textcolor{black}{It also surpasses \textsc{SparseMM}, which adopt unified pruning with head-level budget allocation, indicating the necessity of analyzing head properties and classifying heads accordingly in \method's design.}
These results highlight the advantage of our hybrid compression scheme, which integrates tailored strategies with adaptive budget allocation for different types of attention heads in MLLMs.

\paragraph{Video Tasks.}
We further evaluate \method on three video tasks, with results summarized in \cref{tab:main_result_video}. 
Compared with image input, video input contains temporal information between frames, which poses more challenges to the design of KV cache compression.
Even in this difficult setting, \method achieves higher accuracy across most metrics, while baseline methods struggle to deliver consistent performance. For instance, \textsc{SparseMM} performs well on certain benchmarks (\textsf{NextQA}) but suffers notable degradation on others (5\% average accuracy drop in \textsf{VATEX}) since it only captures head properties in an offline manner. In contrast, \method even exceeds the \textsc{Full Cache} baseline in \textsf{VATEX} by keeping the most salient subset of tokens during decoding, not only preserving but also enhancing the model's capabilities.
This demonstrates the effectiveness of our context-aware design in retaining important information within KV caches for different tasks.

\begin{table*}[!t]
\centering
\renewcommand{\arraystretch}{0.95} 
\scriptsize
\setlength{\tabcolsep}{5pt}
\begin{tabular}{@{}c|l|ccccccccc}
\toprule 
\multicolumn{2}{c|}{\textbf{Method}} & \textbf{\textsf{CL-CH}} & \textbf{\textsf{DocVQA}} & \textbf{\textsf{MMCoQA}} & \textbf{\textsf{MM-QA}} & \textbf{\textsf{SlideVQA}} & \textbf{\textsf{STD}} & \textbf{\textsf{WebQA}} & \textbf{\textsf{WikiVQA}} & \textbf{Average}
\\
\midrule
\multirow{6}{*}{\rotatebox{90}{\textbf{\textit{{\texttt{Qwen2.5-VL-7B}}}}}}
 & 
 \gray{\textsc{Full Cache}} & 
 \gray{43.07} & 
 \gray{97.50} & 
 \gray{66.50} & 
 \gray{75.00} & 
 \gray{83.50} & 
 \gray{29.79} & 
 \gray{76.50} & 
 \gray{92.50} & 
 \gray{70.55} \\
 & \textsc{SnapKV} & 37.41 & 97.50 & 54.00 & 73.00 & 83.00 & 29.02 & 71.00 & 89.50 & 66.80 \\
 & \textsc{LOOK-M} & 38.81 & 97.50 & 52.50 & 73.50 & 82.50 & 29.25 & 71.00 & 88.00 & 66.63 \\
 & \textsc{MadaKV} & 37.86 & 97.50 & 55.00 & 74.50 & 82.00 & 28.24 & 70.50 & 90.00 & 66.95 \\
 & \textsc{SparseMM} & 37.63 & 97.50 & 62.00 & 75.50 & 83.50 & 28.69 & 70.50 & 92.50 & 68.48 \\
 & \method & \textbf{40.57} & \textbf{97.50} & \textbf{63.00} & \textbf{76.00} &\textbf{83.50} & \textbf{29.84} & \textbf{76.00} & \textbf{93.00} & \textbf{69.93} \\
\midrule
\multirow{6}{*}{\rotatebox{90}{\textbf{\textit{{\texttt{Qwen2.5-VL-3B}}}}}}
 & 
 \gray{\textsc{Full Cache}} & 
 \gray{38.34} & 
 \gray{96.00} & 
 \gray{56.50} & 
 \gray{80.00} & 
 \gray{80.00} & 
 \gray{29.37} & 
 \gray{72.00} & 
 \gray{88.00} & 
 \gray{67.53} \\
 & \textsc{SnapKV} & 34.78 & 96.00 & 43.00 & 78.50 & 80.00 & 26.16 & 68.00 & 83.50 & 63.74 \\
 & \textsc{LOOK-M} & 35.12 & 96.00 & 42.50 & 79.00 & 80.00 & 26.24 & 66.50 & 83.50 & 63.61 \\
 & \textsc{MadaKV} & 33.66 & 96.00 & 43.50 & 79.50 & 80.00 & 26.02 & 66.00 & 84.00 & 63.59 \\
 & \textsc{SparseMM} & 35.05 & 96.00 & 50.50 & 79.00 & 79.50 & 26.80 & 65.00 & 87.00 & 64.86 \\
 & \method & \textbf{37.75} & \textbf{96.00} & \textbf{51.00} & \textbf{79.50} & \textbf{80.00} & \textbf{29.22} & \textbf{72.00} & \textbf{88.00} & \textbf{66.68} \\
\midrule
\multirow{6}{*}{\rotatebox{90}{\textbf{\textit{{\texttt{LLaVA-OV-7B}}}}}}
 & 
 \gray{\textsc{Full Cache}} & 
 \gray{44.76} & 
 \gray{97.00} & 
 \gray{53.50} & 
 \gray{77.00} & 
 \gray{77.50} & 
 \gray{31.23} & 
 \gray{76.50} & 
 \gray{88.00} & 
 \gray{68.18} \\
 & \textsc{SnapKV} & 44.13 & 97.00 & 51.50 & 77.00 & 77.50 & 30.17 & 76.50 & 87.50 & 67.66 \\
 & \textsc{LOOK-M} & 44.15 & 97.00 & 51.00 & 77.00 & 77.50 & 30.57 & 76.50 & 87.50 & 67.65 \\
 & \textsc{MadaKV} & 43.37 & 97.00 & 52.00 & 77.00 & 77.50 & 30.69 & 76.50 & 87.50 & 67.70 \\
 & \textsc{SparseMM} & 43.72 & 97.00 & 52.50 & 77.00 & 77.50 & 30.56 & 76.50 & 86.50 & 67.66 \\
 & \method & \textbf{44.31} & \textbf{97.00} & \textbf{52.50} & \textbf{77.00} & \textbf{77.50} & \textbf{30.71} & \textbf{76.50} & \textbf{88.00} & \textbf{67.94}  \\
\bottomrule
\end{tabular}
\caption{Performance of five KV cache compression strategies on three MLLMs on image tasks. 
ROUGE-L~\cite{lin-2004-rouge} evaluates generated–reference consistency for \textsf{CL-CH} and \textsf{STD}, 
while \emph{exact match accuracy} is used for other QA tasks. 
Higher values indicate better performance, and the best measures are highlighted in \textbf{bold} by default. }
\label{tab:main_result_image}
\end{table*}

\begin{table*}[!t]
\renewcommand{\arraystretch}{0.95} 
\scriptsize
\centering
\setlength{\tabcolsep}{3.7pt}
\begin{tabular}{@{}cl|ccccc|c|cccccc}
\toprule

\multicolumn{2}{c|}{\multirow{2}{*}{\textbf{Method}}} & \multicolumn{5}{c|}{\textbf{\textsf{VATEX}}} & \multicolumn{1}{c|}{\textbf{\textsf{NextQA}}} & \multicolumn{6}{c}{\textbf{\textsf{Video-ChatGPT}}} \\

\cmidrule(lr){3-14}

& & \textbf{BLEU-4} & \textbf{Meteor} & \textbf{ROUGE-L} & \textbf{CIDEr} & \textbf{Average} & \textbf{WUPS} & \textbf{CI} & \textbf{DO} & \textbf{CU} & \textbf{TU} & \textbf{CO} & \textbf{Average} \\
\midrule

\multicolumn{1}{c|}{\multirow{6}{*}{\rotatebox{90}
{\textbf{\textit{{\texttt{Qwen2.5-VL-7B}}}}}}}
&
\gray{\textsc{Full Cache}} & 
 \gray{0.2181} & 
 \gray{0.2209} & 
 \gray{0.4266} & 
 \gray{0.4628} & 
 \gray{0.3321} & 
 \gray{33.79} & 
 \gray{3.06} & 
 \gray{3.15} & 
 \gray{3.52} & 
 \gray{2.24} &
 \gray{2.69} & \gray{2.93} \\
\multicolumn{1}{l|}{} & \textsc{SnapKV} & 0.2146 & 0.2125 & 0.4240 & 0.4498 & 0.3252 & 33.29 & 2.90 & 2.96 & 3.40 & 1.95 & 2.53 & 2.75 \\
\multicolumn{1}{l|}{} & \textsc{LOOK-M} & 0.2180 & 0.2174 & 0.4280 & 0.4571 & 0.3301 & 33.07 & 2.92 & 2.97 & 3.38 & 2.05 & 2.49 & 2.76 \\
\multicolumn{1}{l|}{} & \textsc{MadaKV} & 0.2051 & 0.2157 & 0.4228 & 0.4306 & 0.3186 & 33.25 & 2.94 & 3.03 & 3.41 & 2.02 & 2.56 & 2.79 \\
\multicolumn{1}{l|}{} & \textsc{SparseMM} & 0.2133 & 0.2116 & 0.4227 & 0.4198 & 0.3169 & 33.84 & 2.90 & 2.95 & 3.37 & 1.99 & 2.52 & 2.75 \\
\multicolumn{1}{l|}{} & \method & \textbf{0.2266} & \textbf{0.2236} & \textbf{0.4318} & \textbf{0.4774} & \textbf{0.3399} & \textbf{33.86} & \textbf{2.99} & \textbf{3.05} & \textbf{3.47} & \textbf{2.15} & \textbf{2.57} & \textbf{2.85} \\
\midrule
\multicolumn{1}{l|}{\multirow{6}{*}{\rotatebox{90}
{\textbf{\textit{{\texttt{Qwen2.5-VL-3B}}}}}}}
&
 \gray{\textsc{Full Cache}} & 
 \gray{0.2979} & 
 \gray{0.2330} & 
 \gray{0.4896} & 
 \gray{0.5661} & 
 \gray{0.3967} & 
 \gray{29.81} & 
 \gray{2.59} & 
 \gray{2.67} & 
 \gray{3.06} & 
 \gray{1.93} &
 \gray{2.49} & \gray{2.55} \\
\multicolumn{1}{l|}{} & \textsc{SnapKV} & 0.2742 & 0.2186 & 0.4748 & 0.5072 & 0.3687 & 30.10 & 2.64 & 2.67 & 3.09 & 1.77 & 2.23 & 2.48 \\
\multicolumn{1}{l|}{} & \textsc{LOOK-M} & 0.2851 & 0.2245 & 0.4756 & 0.5176 & 0.3757 & 30.49 & \textbf{2.74} & 2.68 & 3.10 & \textbf{1.85} & 2.23 & 2.52 \\
\multicolumn{1}{l|}{} & \textsc{MadaKV} & 0.2691 & 0.2095 & 0.4597 & 0.4708 & 0.3523 & 30.00 & 2.60 & 2.64 & 3.03 & 1.81 & 2.39 & 2.49 \\
\multicolumn{1}{l|}{} & \textsc{SparseMM} & 0.2708 & 0.2080 & 0.4544 & 0.4352 & 0.3421 & \textbf{31.07} & 2.69 & 2.66 & 3.11 & 1.82 & 2.25 & 2.51 \\
\multicolumn{1}{l|}{} & \method & \textbf{0.2936} & \textbf{0.2286} & \textbf{0.4772} & \textbf{0.5427} & \textbf{0.3855} & 30.70 & 2.70 & \textbf{2.72} & \textbf{3.11} & 1.84 & \textbf{2.43} & \textbf{2.56} \\
\midrule
\multicolumn{1}{l|}{\multirow{6}{*}{\rotatebox{90}
{\textbf{\textit{{\texttt{LLaVA-OV-7B}}}}}}}
&
\gray{\textsc{Full Cache}} &
\gray{0.1330} &
\gray{0.1900} &
\gray{0.3739} &
\gray{0.2355} &
\gray{0.2331} &
\gray{20.18} &
\gray{2.99} &
\gray{2.86} &
\gray{3.36} &
\gray{2.00} &
\gray{2.99} &
\gray{2.84} \\
\multicolumn{1}{l|}{} & \textsc{SnapKV} & 0.1161 & 0.1704 & 0.3584 & 0.1799 & 0.2062 & 18.56 & 2.88 & 2.78 & 3.31 & 1.83 & 2.80 & 2.72 \\
\multicolumn{1}{l|}{} & \textsc{LOOK-M} & 0.1205 & 0.1772 & 0.3625 & 0.1957 & 0.2140 & 18.69 & 2.90 & 2.75 & 3.32 & 1.87 & 2.87 & 2.74 \\
\multicolumn{1}{l|}{} & \textsc{MadaKV} & 0.1115 & 0.1748 & 0.3601 & 0.1744 & 0.2052 & 18.79 & 2.90 & 2.79 & 3.35 & 1.90 & 2.89 & 2.77 \\
\multicolumn{1}{l|}{} & \textsc{SparseMM} & 0.1193 & 0.1688 & 0.3593 & 0.1770 & 0.2061 & 18.86 & 2.87 & 2.73 & 3.27 & \textbf{1.91} & 2.84 & 2.72 \\
\multicolumn{1}{l|}{} & \method & \textbf{0.1374} & \textbf{0.1800} & \textbf{0.3700} & \textbf{0.2109} & \textbf{0.2246} & \textbf{19.62} & \textbf{2.96} & \textbf{2.88} & \textbf{3.35} & 1.89 & \textbf{2.89} & \textbf{2.79} \\
\bottomrule
\end{tabular}
\caption{Performance of five KV cache compression strategies on three MLLMs on video tasks. Evaluation metrics (e.g., BLEU-4 and Meteor) are explained in~\cref{appendix:exp_setting}; generally, higher measures indicate better performance. 
}
\label{tab:main_result_video}
\end{table*}

\subsection{Efficiency Results}
\label{exps:efficiency}
%


\begin{table}[!t]
\centering
\footnotesize
\setlength{\tabcolsep}{3.5pt}
\scalebox{0.9}{
\begin{tabular}{lcccc}
\toprule 
\multirow{2}{*}{Method} &
\multirow{2}{*}{Budget} &
Accuracy &
GPU Memory & Latency \\
& & (avg.) & (GB) & (ms/token) \\
\midrule
\gray{\textsc{Full Cache}} & 
\gray{100\%} & \gray{2.93} &
\gray{1.73} & \gray{58.94}
 \\ 
\method & 20\% & \textbf{2.88} & 0.40 & 42.08 \\
\method & 10\% & 2.85 &  \textbf{0.22} & \textbf{38.65} \\
\bottomrule
\end{tabular}
}
\caption{KV cache GPU memory usage and decoding latency on \texttt{Qwen2.5-VL-7B} with \textsf{Video-ChatGPT}. 
}
\label{tab:efficiency_videochatgpt}
\end{table}



We assess the efficiency of \method on \texttt{Qwen2.5-VL-7B} using \textsf{Video-ChatGPT}~\cite{maaz2024video} for real-world long video understanding scenarios. 
We randomly sample 20 data entries and set the maximum generation length to 128 tokens for evaluation.
All experiments use FlashAttention~\cite{dao2023flashattention}. \cref{tab:efficiency_videochatgpt} shows \method markedly reduces both GPU memory and decoding latency relative to the \textsc{Full Cache} baseline. With a 20\% cache budget,
\textcolor{black}{\method results in $4.3\times$ KV cache memory savings and faster decoding;}
at 10\%, 
latency further decreases to 38.65 ms-per-token with only 0.22 GB memory, 
achieving $1.52\times$ decoding speedup and $7.9\times$ KV cache GPU memory reduction compared to \textsc{Full Cache}. 
These efficiency gains show that \method is well-suited for scaling MLLMs to long video and multi-image inputs.
A lower budget enables higher memory utilization and faster decoding. Both 10\% and 20\% are effective compression ratios, with 10\% offering greater deployment flexibility when a minor quality trade-off is acceptable.

\begin{figure} [!t]
\centering
    \includegraphics[scale=0.25]{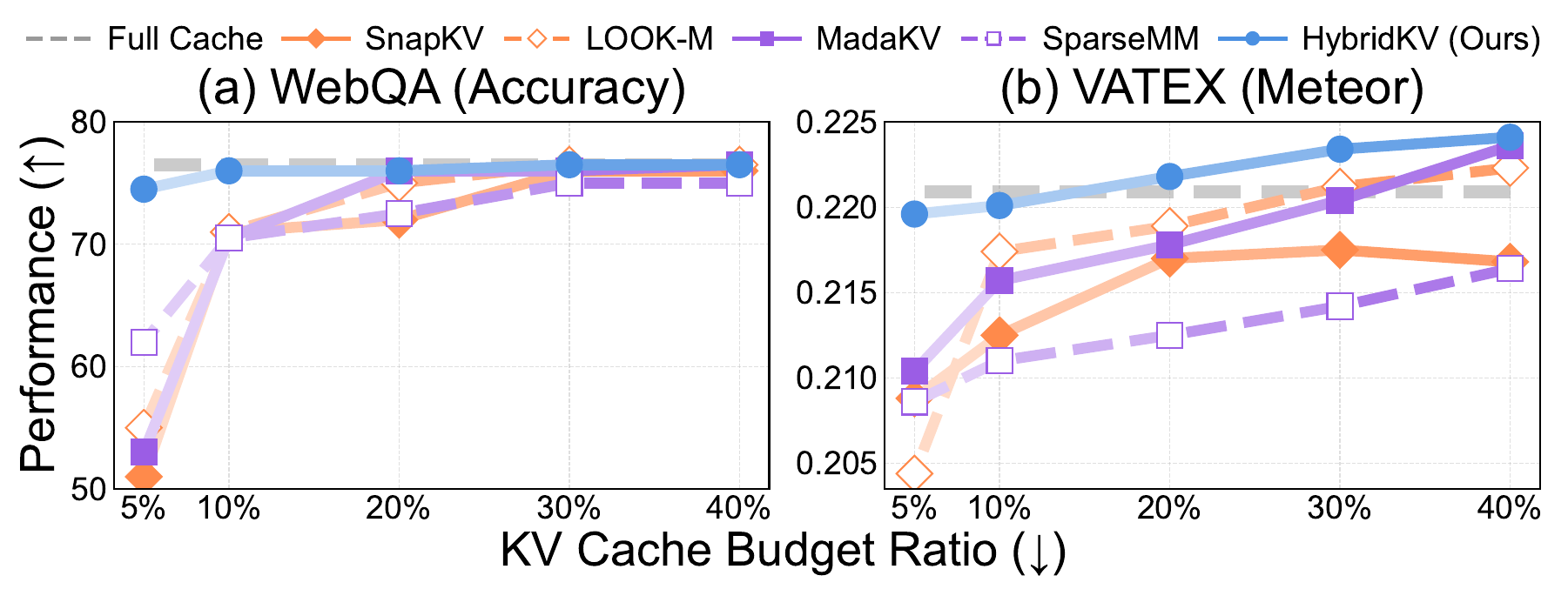}
    \caption{Scalability under different KV cache budgets on \textsf{WebQA} and \textsf{VATEX}.
    \method outperforms all baselines across budgets, matching or exceeding full cache performance and maintaining high accuracy even under aggressive compression (e.g., 5\% budget).
    }
    \label{fig:ablation_compress_ratio}
    \vspace{-1.0cm}
\end{figure}

\subsection{Scaling Law for KV Cache Budgets}

To investigate the scalability of KV cache compression strategies, we conduct experiments on \texttt{Qwen2.5-VL-7B} with cache budgets ranging from 5\% to 40\% on \textsf{WebQA} and \textsf{VATEX}. This setting covers both multi-image QA and video captioning tasks. 
As shown in \cref{fig:ablation_compress_ratio}, \method consistently outperforms all baselines across all budget ratios. Its performance closely tracks the \textsc{Full Cache} baseline and even surpasses it on \textsf{VATEX}, indicating that \method effectively filters out noisy tokens and sharpens focus on salient visual regions. The advantage is most pronounced under extreme compression (<10\% budget), where baselines degrade sharply but \method maintains strong accuracy. 
These results highlight that \method scales effectively by preserving critical information in KV caches, thereby reducing memory cost with minimal performance loss.

\subsection{Ablation Study}

Here, we analyze the effect of each component in \method through ablations on head classification and budget allocation. Results are reported on three subtasks, Spot-the-Diff (\textsf{STD}), \textsf{WebQA} for image tasks, and \textsf{NextQA} for video tasks, with latency measured under a 10\% KV cache budget.

\paragraph{Effect of Head Classification.}

\cref{tab:ablation_head_class} compares \method with two variants that assign all heads as either static or dynamic under the same budget allocation strategy. Both variants suffer accuracy degradation across tasks, since treating all heads identically overlooks their heterogeneous patterns in MLLMs. 
The variant with all dynamic heads results in higher latency due to more KV cache I/O overhead.
In contrast, \method leverages text-centric classification to adaptively identify head properties from input contexts, yielding better accuracy without significant latency overhead.

\begin{table}[t]
\setlength{\tabcolsep}{2pt} 
    \centering
    \small
     \begin{tabular}{lcccc}
    \toprule
        \multirow{2}{*}{Method} & \multirow{2}{*}{\textsf{STD}} & \multirow{2}{*}{\textsf{WebQA}} & \multirow{2}{*}{\textsf{NextQA}} & Latency \\
        & & & & (ms/token) \\ 
    \midrule
        \method 
        & \textbf{29.84} 
        & \textbf{76.00} 
        & \textbf{33.86}
        & {38.65} \\
        
        w. all static heads 
        & 28.72
        & 73.00 
        & 33.61 
        & \textbf{34.13} \\
        
        w. all dynamic heads 
        & 29.77
        & 75.50 
        & 33.76
        & 48.15 \\
    \bottomrule
    \end{tabular}
    \caption{Effect of head classification. }
    \label{tab:ablation_head_class}
\end{table}

\paragraph{Effect of Budget Allocation.}

\cref{tab:ablation_budget_alloc} compares \method with two variants fixing the same head classification process: one that removes head-level adaptive allocation for static heads, and a uniform baseline that ignores both head type and individual head. 
\method exhibits better accuracy on both image and video tasks while maintaining decoding latency, showing the robustness of our budget allocation. Removing head-level allocation (i.e.~w/o H) degrades accuracy due to ignoring the variable importance of individual heads. 
When head-type allocation is also removed (i.e.~w/o (H \& HT)), accuracy declines further because tailored strategies for static and dynamic heads are ignored, leading to imbalanced budgets. This also increases decoding latency due to the over-allocation to dynamic heads.
These results confirm that our top-down allocation, combining head-type and head-level granularity, improves both accuracy and efficiency and complements the hybrid compression design.

\begin{table}[t]
\setlength{\tabcolsep}{4pt} 
    \centering
    \small
    \begin{tabular}{lcccc}
    \toprule
        \multirow{2}{*}{Method} & \multirow{2}{*}{\textsf{STD}} & \multirow{2}{*}{ \textsf{WebQA}} & \multirow{2}{*}{\textsf{NextQA}} & Latency \\
        & & & & (ms/token) \\ 
    \midrule
        \method & \textbf{29.84} & \textbf{76.00}  & \textbf{33.86} & \textbf{38.65} \\
        w/o H & 29.67 & 76.00 & 33.70 & 38.65 \\
        w/o (H \& HT) & 29.58 & 74.50 & 33.29 & 44.67 \\
    \bottomrule
    \end{tabular}
    \caption{Effect of budget allocation. H: head-level budget allocation; HT: head-type-level budget allocation.}
    \label{tab:ablation_budget_alloc}
\end{table}

\section{Related Work}
Existing KV cache compression methods in LLMs fall into \emph{static pruning} and \emph{dynamic retrieval}. Static methods~\cite{xiao2024efficient, zhang2023h2o, li2024snapkv} discard KV caches in the prefill stage using fixed criteria, while dynamic methods~\cite{tang2024quest, sun2024shadowkv, chen2024arkvale} retrieve important KV caches during decoding. These paradigms extend to MLLMs with compression strategies at the token-level~\cite{wan2024look}, layer-level~\cite{li2025madakv, wan2025meda}, and head-level~\cite{wang2025sparsemm}.
However, a common limitation is that they apply either a fully static or a fully dynamic strategy, thus overlooking the diverse patterns of attention heads. In contrast, \method is founded on the observation of heterogeneous attention patterns and introduces a novel hybrid KV cache compression framework for efficient MLLM inference. 
We classify heads into static and dynamic types using text-centric attention embedded in multimodal inputs, and apply a tailored budget allocation and compression strategy to each.

Another line of work directly addresses the redundancy of visual information by pruning or merging visual tokens, either in the vision encoder~\cite{yang2025visionzip} or within the language model~\cite{chen2024image, zhangsparsevlm, hu2025lightvlm}. These methods rely on attention scores or frame uniqueness~\cite{liu2025video} to identify and discard less important tokens. Differently, our method tackles visual redundancy from the heterogeneity of attention heads in MLLMs, providing a more fundamental and effective mechanism for managing visual information within the KV cache. \cref{appendix:related_work} provides an extended discussion of related work.

\section{Conclusion}
We present \method, a hybrid KV cache compression framework for efficient MLLM inference. 
By identifying distinct attention head patterns and applying hierarchical, head‑aware compression, \method cuts KV cache GPU memory by $7.9\times$ and speeds up decoding by $1.52\times$, all while preserving generation quality on \texttt{Qwen2.5-VL-7B}. 
We envision \method as both a valuable community tool for deployment‑ready MLLMs and a foundation for future strategy‑driven, context‑aware compression in multimodal reasoning.

\clearpage
\section{Ethical Considerations}
All experiments in this work are conducted using open-source datasets and models. Our research focuses solely on improving inference efficiency and does not involve any sensitive data, human subjects, or commercial use~\footnote{\url{https://github.com/MileBench/MileBench} (Apache License 2.0)
}
\footnote{\url{https://huggingface.co/datasets/lmms-lab/VATEX} (Apache License 2.0)
}
\footnote{\url{https://huggingface.co/datasets/lmms-lab/NExTQA} (Apache License 2.0)
}
\footnote{\url{https://huggingface.co/datasets/lmms-lab/VideoChatGPT} (Apache License 2.0)
}.

\section{Limitation}
While \method delivers strong improvements in memory efficiency and decoding speed, it also has certain limitations.  
First, our head classification relies on text-centric attention patterns observed during the prefill stage. Although we show this to be a reliable proxy, the classification depends on thresholding and may vary across model scales or domains. More sophisticated classifiers (e.g., learned or adaptive) could further improve robustness but are beyond the focus of this work.  
Second, the current design targets inference-time efficiency without involving additional training or fine-tuning. This makes \method broadly applicable, but it also limits opportunities to co-adapt compression strategies with model training.  
Third, our evaluation primarily considers image and video benchmarks where KV cache growth is most severe. Extending to other modalities and tasks (e.g., audio-grounded MLLMs or extremely long-text reasoning) could reveal additional challenges.  

Nevertheless, \method is compatible with future extensions such as adaptive thresholding, joint training of head classification, or integration with other compression paradigms (e.g., quantization or speculative decoding). We believe these directions will further enhance the generality and scalability of \method.

\bibliography{custom}

\clearpage

\appendix

\section{Details of Hybrid KV Cache Compression}
\label{appendix:hybrid_kv_compress_impl}

\begin{figure} [!t] 
\centering
    \includegraphics[scale=0.34]{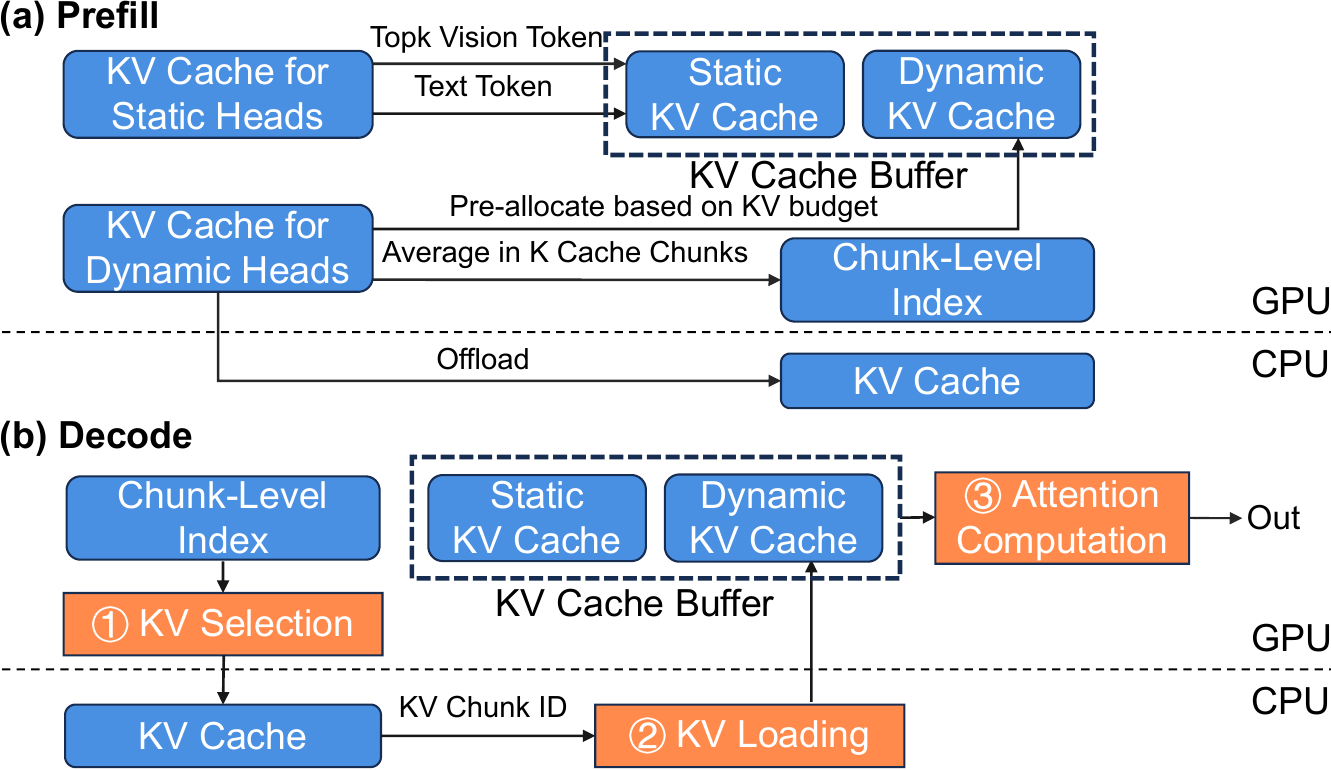}
    \caption{
        Illustration of hybrid KV cache compression.
    }
    \label{fig:kv_pipeline_map}
\end{figure}

A detailed description of our hybrid KV cache compression implementation is provided in~\cref{fig:kv_pipeline_map}.

\paragraph{KV Cache Pruning for Static Heads.} For static heads, 
we apply a selective pruning strategy to fit their allocated budget. Inspired by methods like SparseMM~\cite{wang2025sparsemm}, our policy employs an observation window to efficiently score the importance of all tokens within the initial prompt. 

The final condensed cache is constructed from three distinct sets of tokens: (i) the observation window itself, (ii) all text tokens from the historical context (tokens before the window), and (iii) the Top-$M$ most relevant visual tokens from the historical context.
To determine the relevance of historical visual tokens, we use the queries from the observation window to attend to the entire prompt. Given query states $\mathbf{Q}$ and key states $\mathbf{K}$, we compute vector $\mathbf{s}_w$ by the attention scores:
\begin{equation}
\mathbf{S}_{w} = \text{softmax}\left(\frac{\mathbf{Q}[C-w:C, :] \mathbf{K}^\top}{\sqrt{d_k}}\right)
\end{equation}
\begin{equation}
\mathbf{s}_w = \frac{1}{w} \sum_{i=1}^{w} \mathbf{S}_{w}[i,:]
\end{equation}
where $w$ denotes the length of the observation window, and $C$ is the length of the entire context. This vector $\mathbf{s}_w \in \mathbb{R}^{C}$ assigns a relevance score to every token in the prompt. We then select the Top-$M$ historical visual tokens based on their corresponding scores in this vector. The number of visual tokens to keep, $M$, is set by the remaining budget after accounting for the observation window and historical text tokens. This policy guarantees the preservation of the most recent context, complete textual instruction, and the most significant historical visual features.

\paragraph{KV Cache Retrieval for Dynamic Heads.}
For dynamic heads, we adopt a chunk-wise retrieval strategy widely used in recent KV cache methods~\cite{tang2024quest,sun2024shadowkv}. Specifically, the token sequence is divided into fixed-size chunks (e.g., 8 tokens each), and each chunk is represented by metadata computed as the average of its key vectors. As illustrated in \cref{fig:kv_pipeline_map}, during the prefill stage we build a chunk-level index from the metadata, offload dynamic-head KV caches to CPU memory, and retain only the index on GPU. At each decoding step, we identify relevant chunks by computing the inner product between the query vector and chunk metadata, load the selected chunks from CPU, and update the GPU KV buffer accordingly. The number of loaded chunks is determined by the per-head budget and chunk size. The final attention output is then computed together with the static KV cache. This strategy adaptively retrieves essential contexts while achieving efficiency through chunk-wise data transfer.

\section{Details of Related Work}
\label{appendix:related_work}
\paragraph{KV Cache Compression.} 
In the LLM era, these methods fall into two categories: static pruning and dynamic retrieval. For static pruning methods, 
KV caches are discarded in the prefill stage. StreamingLLM~\cite{xiao2024efficient} retains attention sinks and recent tokens, while H2O~\cite{zhang2023h2o} and SnapKV~\cite{li2024snapkv} retain important tokens using cumulative attention scores. PyramidKV~\cite{cai2024pyramidkv} allocates KV cache budgets across layers based on informativeness. Some methods focus on the head-level properties, DuoAttention~\cite{xiaoduoattention} and RazorAttention~\cite{tang2025razorattention} divide attention heads into retrieval and non-retrieval categories, and store all tokens in retrieval heads while applying pruning for others. AdaKV~\cite{feng2024ada}, on the other hand, proposes an adaptive budget allocation algorithm and can be integrated into existing methods. However, these one-time static pruning methods perform well in efficiency but often fail in model performance due to severe information loss. 
For dynamic retrieval methods, in the prefill stage, total KV caches are stored in the GPU memory~\cite{tang2024quest}, or offloaded to CPU to reduce memory overhead~\cite{sun2024shadowkv,chen2024arkvale}, and important KV caches are retrieved during the decode stage. Quest~\cite{tang2024quest}, ShadowKV~\cite{sun2024shadowkv} and ArkVale~\cite{chen2024arkvale} summarize the Key cache chunks into compact representations with different ways (e.g.,~mean-pooling) and retrieve important KV caches in a chunk-wise manner. MagicPig~\cite{chenmagicpig} applies locality-sensitive hashing (LSH) to sample and retrieve significant tokens. These methods achieve better model performance at the cost of complexity and efficiency. \method uses the basic strategy of chunk-wise retrieval and is orthogonal to the specific optimizations of these methods.
The above methods for LLMs focus on text-based KV cache compression while overlooking multimodal contexts. They are not directly applicable to multimodal models due to the unique characteristics of cross-modal attention.

As multimodal understanding scenarios progress, KV cache compression methods are further designed for MLLMs and can be divided into three categories: token-level compression, layer-level compression, and head-level compression. For token-level methods, LOOK-M~\cite{wan2024look} prioritizes the retention of text tokens over visual tokens for modality-aware compression. For layer-level methods, MadaKV~\cite{li2025madakv} proposes modality preference metric and applies hierarchical compression for attention layers. MEDA~\cite{wan2025meda} leverages cross-modal attention entropy for dynamic KV cache allocation across layers. For head-level methods, SparseMM~\cite{wang2025sparsemm} observes the visual sparsity of attention heads in MLLMs and allocates asymmetric KV cache budgets to heads based on their visual scores.
However, the above methods, whether designed for LLMs or MLLMs, treat all the attention heads either in a static manner or in a dynamic manner, leading to suboptimal compression results. 
In contrast, \method is tailored for optimizing MLLMs and observes the heterogeneous attention patterns across heads. It designs a novel hybrid KV cache compression framework for MLLMs to achieve better model performance with significant efficiency gains.

\paragraph{Visual Token Compression for MLLMs.}
Existing research proposes various visual token compression methods due to the great redundancy of visual tokens. FastV~\cite{chen2024image} prunes visual tokens after Layer 2 using attention scores. SparseVLM~\cite{zhangsparsevlm} uses text-to-vision attention for scoring to evict unimportant visual tokens. VisionZip~\cite{yang2025visionzip} reduces visual redundancy in the vision encoders. LightVLM~\cite{hu2025lightvlm} proposes pyramid token merging in the prefill stage and KV cache compression in the decode stage. There also exist methods that focus on video inputs because of the high volume of video tokens. Dycoke~\cite{tao2025dycoke} applies token merging across video frames and reduces KV cache dynamically. VidCom$^2$~\cite{liu2025video} dynamically adjusts compression intensity based on frame uniqueness. Differently, our method operates on another perspective that leverages the heterogeneous attention head patterns embedded in MLLMs to address the visual redundancy issues.

\section{Details of Benchmark}
\label{appendix:exp_setting}


To comprehensively evaluate our model, we utilize a series of established benchmarks for both image and video understanding.
For image tasks, we select a multifaceted suited of tasks from MileBench. The selection includes visual difference caption on the CLEVR-Change (\textsf{CL-CH})~\cite{park2019robust} and Spot-the-Diff (\textsf{STD})~\cite{jhamtani2018learning} datasets, measured by ROUGE-L (Recall-Oriented Understudy for Gisting Evaluation-Longest Common Subsequence)~\cite{lin-2004-rouge}. It also incorporates a diverse set of visual question answering (VQA) tasks that span various domains: document and presentation understanding (\textsf{DocVQA}~\cite{mathew2021docvqa}, \textsf{SlideVQA}~\cite{tanaka2023slidevqa}), web-based knowledge extrantion (\textsf{WebQA}~\cite{chang2022webqa}, \textsf{WikiVQA}~\cite{yang2015wikiqa}), and joint reasoning over text, tables, and images (\textsf{MMCoQA}~\cite{li2022mmcoqa}, MultiModalQA (\textsf{MM-QA})~\cite{talmor2021multimodalqa}. Performance across all these VQA tasks is measured by accuracy. 
For video tasks, we employ \textsf{VATEX}~\cite{wang2019vatex} for video captioning, with performance measured by four metrics: BLEU-4 (4-gram Bilingual Evaluation Understudy)~\cite{papineni2002bleu}, METEOR (Metric for Evaluation of Translation with Explicit ORdering)~\cite{banerjee2005meteor}, ROUGE-L~\cite{lin-2004-rouge}, and CIDEr (Consensus-based Image Description)~\cite{vedantam2015cider}. For \textsf{NextQA}~\cite{xiao2021next}, which focuses on temporal and causal reasoning, we use the metric Wu-Palmer Similarity (WUPS) score~\cite{malinowski2014multi} to evaluate the quality of the generated answers. Additionally, we leverage the \textsf{Video-ChatGPT}~\cite{maaz2024video} to evaluate conversational abilities across five dimensions: Correctness of Information (CI), Detail Orientation (DO), Contextual Understanding (CU), Temporal Understanding (TU) and Consistency (CO). These metrics are accessed using an LLM-generated prediction score ranging from 0 to 5 (GPT Score) with gpt-4o-mini.


\section{Sensitivity Study on Hyperparameters}
\label{appendix:hyperparam}
We conduct a sensitivity analysis to validate the choices of our key hyperparmeters, score threshold $\theta$ and share coefficient $r$. All experiments are performed on the \texttt{Qwen2.5-VL-7B} model across three datasets: CLEVR-Change (\textsf{CL-CH}), Spot-the-Diff (\textsf{STD}), and \textsf{WebQA}. When analyzing one hyperparameter, all other parameters are held constant at their default optimal values.
\paragraph{Score Threshold $\theta$ in Head Classification.} 
\label{appendix:param:threshold}
The score threshold $\theta$ is a critical hyperparameter in our text-centric head classification process, acting as the decision boundary for categorizing an attention head as static or dynamic. The design of $\theta$ is based on the intrinsic property of attention sparsity, which results in the scores for most heads being clustered at the higher end of the spectrum. Consequently, any meaningful decision boundary for separating static from dynamic heads must also reside in a high-value range. Thus, our search is focused on the high-score interval $[0.75, 0.95]$. 
The results, summarized in Table~\ref{tab:threshold_tau}, show that the choice of $\theta$ directly impacts model performance. A threshold set too low, such as $\theta=0.75$ may incorrectly classify heads with dynamic properties as static. This prevents these heads' recall of critical tokens from the context, leading to a loss of essential information and causing a drop of 1.8 points in average accuracy compared to the peak performance. Conversely, an excessively high threshold like $\theta=0.95$ creates too many dynamic heads. This not only increases data transmission overhead during inference due to the costly retrieval operations, but also heightens the risk of introducing noise from irrelevant tokens, resulting in a slight dip in performance. Therefore, \textbf{we select $\theta=0.9$ as it achieves the best accuracy while maintaining a reasonable number of dynamic heads for efficiency.}

\begin{table}[t]
    \setlength{\tabcolsep}{9pt} 
    \centering
    \small
    \begin{tabular}{lcccc}
    \toprule
        \textbf{$\theta$} & \textsf{CL-CH} & \textsf{STD} & \textsf{WebQA} & Average \\
    \midrule
        0.75 & 39.06 & 29.53 & 72.50 & 47.03 \\
        0.80 & 39.49 & 29.41 & 72.50 & 47.13 \\
        0.85 & 39.87 & 29.51 & 75.00 & 48.13 \\
        \textbf{0.90} & \textbf{40.57} & \textbf{29.84} & \textbf{76.00} & \textbf{48.83} \\
        0.95 & 39.13 & 29.69 & 76.00 & 48.27 \\
    \bottomrule
    \end{tabular}
    \caption{Sensitivity study on the score threshold $\theta$. For this analysis, the share coefficient is fixed at $r=0.75$ and the KV cache budget is 10\%.}
    \label{tab:threshold_tau}
\end{table}

\paragraph{Share Coefficient $r$ in Budget Allocation.}
\label{appendix:param:share_ratio}
In our top-down budget allocation process, the share coefficient $r$ controls the proportion of the KV cache for two types of head. Our search for its optimal value is designed to cover several operational regimes to fully characterize its impact: values of $r\ll 1.0$ to investigate a resource starvation scenario for the dynamic heads, values of $r\approx1.0$ to access a balanced allocation, and values of $r > 1.0$ to observe the effects of favoring dynamic heads. Notably, $r$ is naturally constrained by the total budget and must satisfy the condition $r \le B_{\text{total}} / (\bar{B} * N_{\text{dyna}})$ to avoid exceeding available cache budget. 
The results in Table~\ref{tab:share_ratio_r} clearly illustrate the behavior across these regimes. A coefficient set too low provides insufficient budget for the dynamic heads, harming their ability to retrieve important tokens and degrading model accuracy. Conversely, a coefficient set too high is also suboptimal. It allocates an excessive budget to the dynamic heads, which increase data transmission overhead without any meaningful gain in accuracy, while simultaneously starving the static heads of their required resources. This imbalance ultimately harms collective performance. This demonstrates the need for a moderate, well-balanced coefficient. Therefore, \textbf{we select $r=0.75$ as the optimal choice as it achieves the highest empirical accuracy. This moderate value strikes the ideal balance, providing a relatively sufficient and balanced budget for both types of head, while avoiding the increased data transmission overhead and budget overflow risk associated with higher coefficients.}

\begin{table}[t]
    \setlength{\tabcolsep}{9pt} 
    \centering
    \small
    \begin{tabular}{lcccc}
    \toprule
        \textbf{$r$} & \textsf{CL-CH} & \textsf{STD} & \textsf{WebQA} & Average \\
    \midrule
        0.1 & 39.26 & 29.37 & 75.00 & 47.88 \\
        0.25 & 39.87 & 29.65 & 75.50 & 48.34 \\
        0.50 & 40.03 & 29.75 & 75.50 & 48.43 \\
        \textbf{0.75} & \textbf{40.57} & \textbf{29.84} & \textbf{76.00} & \textbf{48.83} \\
        1.00 & 40.12 & 29.68 & 74.00 & 47.93 \\
        1.25 & 39.02 & 29.44 & 74.50 & 47.65 \\
    \bottomrule
    \end{tabular}
    \caption{Sensitivity study on the share coefficient $r$. For this analysis, the score threshold is fixed at $\theta=0.90$ and the KV cache budget is 10\%.}
    \label{tab:share_ratio_r}
\end{table}

\section{Case Study}
\label{appendix:case_study}

\begin{figure*} [t!] 
\centering
    \includegraphics[scale=0.6]{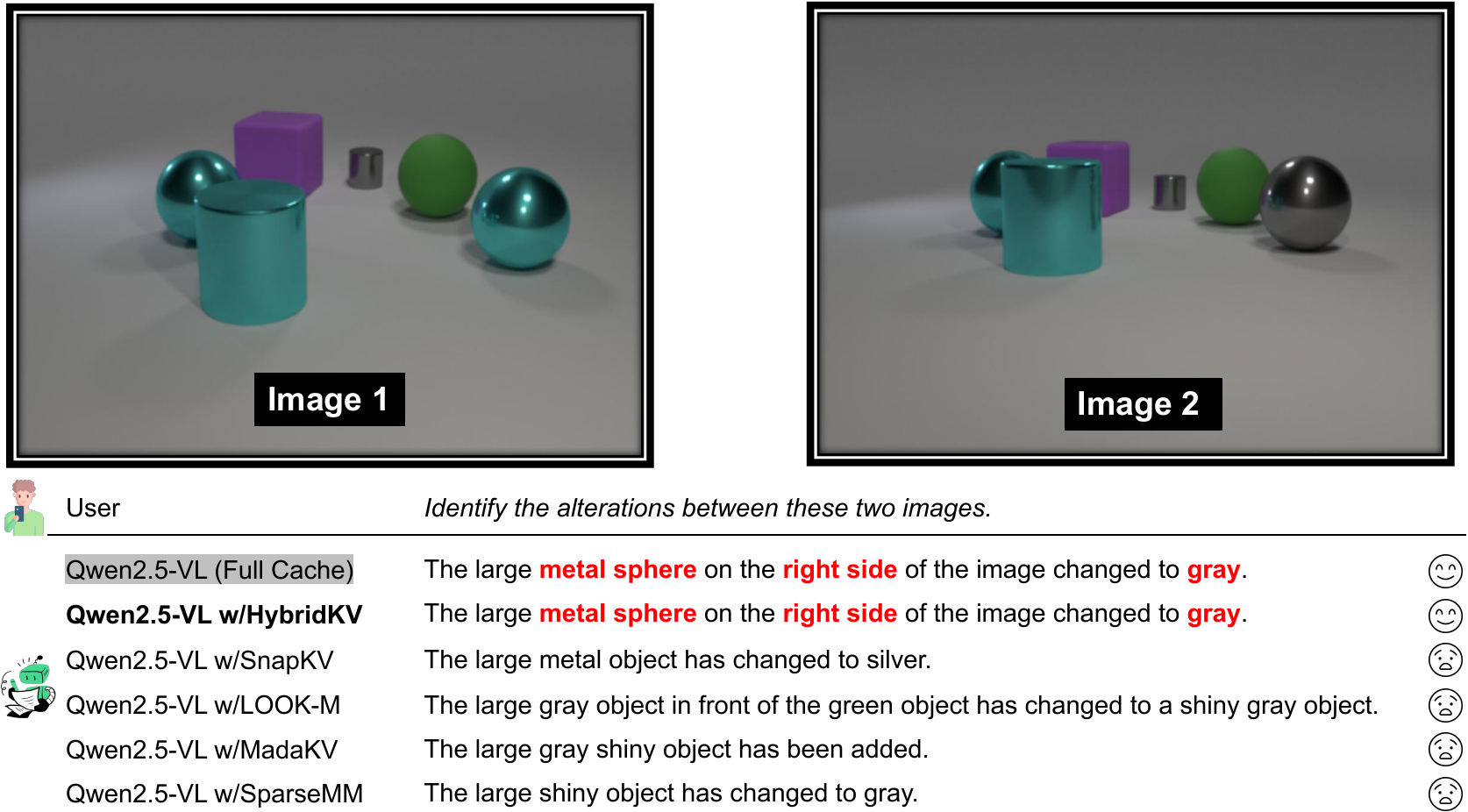}
    \caption{
        Case study of \method on \textsf{CL-CH} using \texttt{Qwen2.5-VL-7B}. Other methods generate the \textit{wrong} answer after KV cache compression, while \method still retains the correct answer.
    }
    \label{fig:appendix:case_study_image}
\end{figure*}

\begin{figure*} [t!] 
\centering
    \includegraphics[scale=0.58]{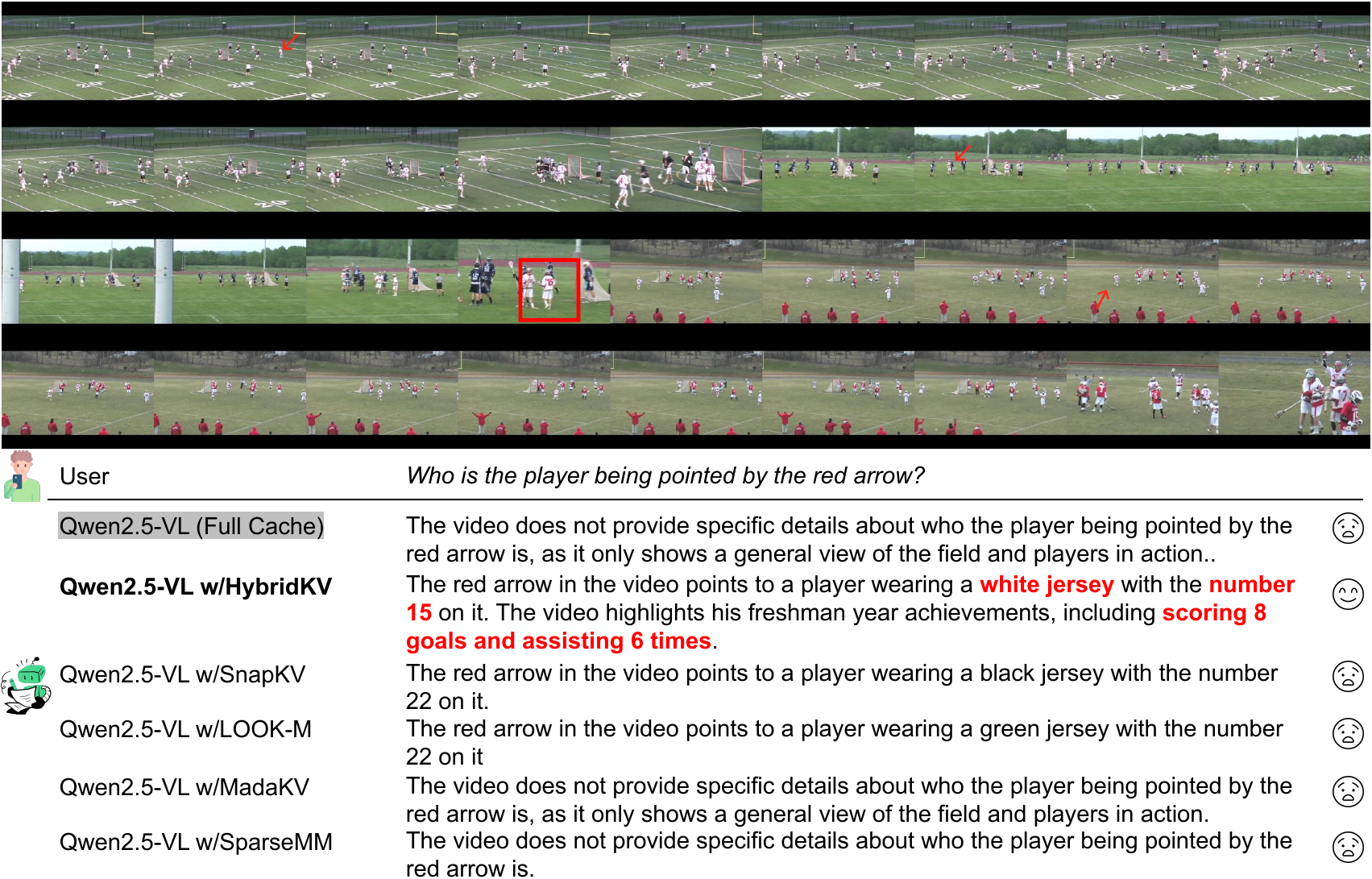}
    \caption{
        Case study of \method on \textsf{Video-ChatGPT} using \texttt{Qwen2.5-VL-7B}. \method enables the decode tokens to adaptively focus on critical visual regions, highlighted by red font and boxes, while the model with full cache attend to the noisy tokens, leading to degraded accuracy.  This demonstrates that retaining a small yet salient subset of tokens during decoding can not only preserve but even enhance the model's capabilities.
    }
    \label{fig:appendix:case_study_video}
\end{figure*}

Two illustrative examples for image and video tasks are presented in~\cref{fig:appendix:case_study_image} and~\cref{fig:appendix:case_study_video}, respectively.
\method achieves superior generation quality by enabling the model to focus on critical visual regions through a head-level hybrid design. This allows it to not only outperform other KV cache compression baselines, but also surpass the model with full cache.

\end{document}